\documentclass[letterpaper, 10 pt, conference]{ieeeconf}  %

\IEEEoverridecommandlockouts                              %

\usepackage{graphics} %
\usepackage{epsfig} %
\usepackage{times} %
\usepackage{amsmath} %
\usepackage{amssymb}  %
\usepackage{subfig}
\usepackage{url}
\usepackage{hyperref}

\title{\LARGE \bf
GRIT: Fast, Interpretable, and Verifiable Goal Recognition with Learned Decision Trees for Autonomous Driving
}

\author{\authorblockN{Cillian Brewitt\authorrefmark{1}, Balint Gyevnar\authorrefmark{1}, Samuel Garcin\authorrefmark{1}, Stefano V. Albrecht\authorrefmark{1}\authorrefmark{2}}%
\authorblockA{{\tt\small \{cillian.brewitt, balint.gyevnar, s.garcin, s.albrecht\}@ed.ac.uk}}
\authorblockA{\authorrefmark{1}School of Informatics, University of Edinburgh, UK}
\authorblockA{\authorrefmark{2}Five\,AI Ltd., UK}
        \thanks{{\bf GRIT code:} \url{https://github.com/uoe-agents/GRIT}}
        \vspace{-1.0em}
        }

\begin{document}

\maketitle
\thispagestyle{empty}
\pagestyle{empty}

\begin{abstract}

It is important for autonomous vehicles to have the ability to infer the goals of other vehicles (goal recognition), in order to safely interact with other vehicles and predict their future trajectories. This is a difficult problem, especially in urban environments with interactions between many vehicles. Goal recognition methods must be fast to run in real time and make accurate inferences. As autonomous driving is safety-critical, it is important to have methods which are human interpretable and for which safety can be formally verified. Existing goal recognition methods for autonomous vehicles fail to satisfy all four objectives of being fast, accurate, interpretable and verifiable. We propose \textit{Goal Recognition with Interpretable Trees} (GRIT), a goal recognition system which achieves these objectives. GRIT makes use of decision trees trained on vehicle trajectory data. We evaluate GRIT on two datasets, showing that GRIT achieved fast inference speed and comparable accuracy to two deep learning baselines, a planning-based goal recognition method, and an ablation of GRIT. We show that the learned trees are human interpretable and demonstrate how properties of GRIT can be formally verified using a satisfiability modulo theories (SMT) solver.

\end{abstract}

\section{INTRODUCTION}
\begin{figure}[t]
    \centering
    \includegraphics[width=0.485\textwidth]{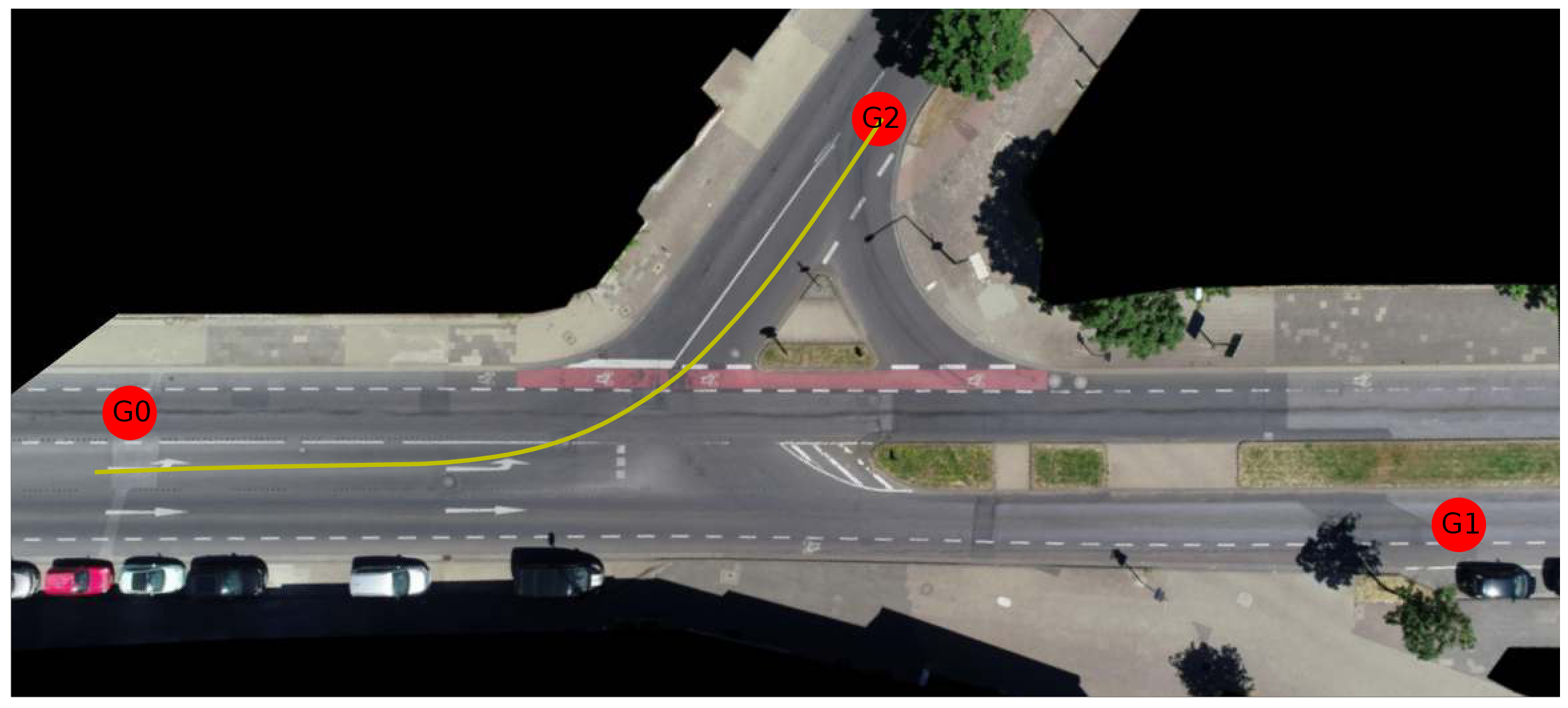}
    \caption{The ``Heckstrasse" junction from the inD dataset \cite{bock_ind_2019}. Goal locations G0, G1 and G2 are shown by red circles. An example vehicle trajectory is shown by the yellow line.}
    \label{fig:heckstrasse}
    \vspace{-1.0em}
\end{figure}

To safely navigate through busy city traffic, autonomous vehicles (AVs) must be able to predict the future trajectories of other road users, and an effective method of doing this is to first recognise their goals. For example, the goal of a vehicle could be to take a certain exit at a junction, as shown in Figure \ref{fig:heckstrasse}. There are several desirable properties for goal recognition methods: these methods must be \textbf{fast}, as AVs must make decisions in real time and quickly react to new information; and predictions must be \textbf{accurate} to be be useful for planning and navigation. It is also desirable for goal recognition methods to be \textbf{interpretable} by humans. Regulations which codify the ``right to an explanation" for some types of automated decision have already been created \cite{goodman_european_2017}, and regulators may create similar rules for AVs.

Prediction accuracy is typically measured empirically based on statistical averages, but no guarantees can be given about inferences made \cite{fisher_towards_2021}. Autonomous driving is a safety-critical task, and it is important to have ways of validating that prediction methods will act as intended when deployed. The amount of data required to empirically validate the safety of an autonomous vehicle is enormous, on the order of billions of miles \cite{shalev-shwartz_formal_2017}. An alternative approach to safety validation is to formally \textbf{verify} models of the system to guarantee safety under all possible conditions \cite{butler_infeasibility_1993,Luckcuck_formal_2019}. 

Verification and interpretation are not possible for some prediction methods such as those that make use of deep neural networks (DNNs) \cite{rhinehart_precog_2019,lee_desire_2017,chai_multipath_2019,xu_learning_2019,casas_intentnet_2018,hu_generic_2019,li_conditional_2019,mozaffari_deep_2020}, due their complexity and large number of parameters \cite{ayers_parot_2020}. Another approach to prediction is to first perform goal recognition, and use this to inform trajectory prediction. One method of goal recognition is planning to a set of possible goal locations from the perspective of the agent for which we are performing goal recognition on \cite{albrecht_interpretable_2021,hardy_contingency_2013,bandyopadhyay_intention-aware_2013,ziebart_maximum_2008,darweesh_estimating_2019,hanna_interpretable_2021}. Such goal recognition methods can be used to generate accurate long term trajectory predictions which are explainable through rationality principles. However, the planning process is computationally complex, making it difficult to run in real time and intractable to verify.

There has been significant previous work on verification of autonomous driving policies \cite{zita_application_2017, shalev-shwartz_formal_2017}. However, to the best of our knowledge no existing verifiable prediction methods for AVs. Recent work has shown that decision trees can produce models which are more easily verified and interpreted than deep neural networks in domains other than prediction for autonomous vehicles. Bastani et al. \cite{bastani_verifiable_2018} used decision trees to represent a reinforcement learning policy, and showed that certain properties of these decision trees can be efficiently verified using off-the-shelf satisfiability modulo theories (SMT) solvers. Liu et al. \cite{liu_improving_2019} used knowledge distillation to obtain an interpretable decision tree from a less interpretable deep neural network for classification. These works motivate our approach to use decision trees for an interpretable and verifiable goal recognition method.

In this paper, we present \textit{Goal Recognition with Interpretable Trees} (GRIT), a vehicle goal recognition method which satisfies the objectives of being fast, accurate, interpretable and verifiable. At the core of this method we use decision trees which are trained from vehicle trajectory data. Decision trees are computationally efficient and highly structured, which allows GRIT to be fast at inference time and interpretable to humans. We show how properties of GRIT can be verified automatically by mapping the learned trees into propositional logic and using an SMT solver \cite{de_moura_z3_2008}. We evaluate GRIT across four scenarios from two vehicle trajectory datasets \cite{bock_ind_2019,krajewski_round_2020} and show that it achieves comparable accuracy to deep learning baselines and performs inference fast enough to run in real time. We demonstrate by example how the trained decision trees are human interpretable. We verified several properties of the models, for example verifying that the probability of a goal is above a threshold if a vehicle is in the correct lane for that goal. If verification fails, the SMT solver provides a counterexample which can teach us about the way in which the model works, facilitating inspection and debugging. To the best of our knowledge, GRIT is the first goal recognition method for autonomous vehicles which has been shown to be verifiable.

\section{Problem Definition}

We consider a set of vehicles $\mathcal{I}$. Each vehicle $i \in \mathcal{I}$ at time $t$ has a state $s^i_t \in \mathcal{S}^i$. The state of a vehicle comprises of its pose (location and orientation), speed and acceleration. The state of all vehicles at time $t$ is given by $s_t = \{s_t^i | i \in \mathcal{I} \} \in \mathcal{S}$. We write $s_{a:b}=\{s_a,s_{a+1},...,s_b\}$ to represent the state of all vehicles between times $a$ and $b$. For each vehicle $i$ at time $t$, we assume a set of possible goals $G^i_t = \{G^{i,1}_t, G^{i,2}_t, ...\}$ with $G_t^{i,k} \subset S^i$. That is, we assume that the goal of a vehicle is to reach a certain state such as a target location. Examples of such goals are exits at a junction or the visible end of a lane. If a vehicle $i$ has trajectory $s^i_{1:n}$ and $s^i_n \in G^{i,k}_t$, then the vehicle has reached goal $G^{i,k}_t$. The static scene information, which in our case is the local road layout, is represented by $\phi$. We define the goal recognition problem as the task of inferring $P(G^{i,k}_t|s_{1:t},\phi)$, a probability distribution over goals for vehicle $i$ at time $t$.

\section{GRIT: Goal Recognition with Interpretable Trees}

Our method aims to infer a probability distribution over goals for a vehicle based on past observations, using models trained from vehicle trajectory data. We use the name GRIT (Goal Recognition with Interpretable Trees) to refer to the entire training and inference process. GRIT computes a Bayesian posterior probability distribution over goals
\begin{equation} \label{goal_prob_overview}
P(G^i|s_{1:t},\phi) = \frac{L(s_{1:t}|G^i,\phi) P(G^i|\phi)}{\sum_{G^\prime \in G^i_t}{L(s_{1:t}|G^\prime,\phi) P(G^\prime|\phi)}}
\end{equation}
and represents the likelihood $L(s_{1:t}|G^i,\phi)$ of a trajectory $s_{1:t}$ given goal $G^i$ using decision trees learned from vehicle trajectory data prior to deployment.

\begin{figure}[t]
    \centering
    \includegraphics[height=0.26\textheight]{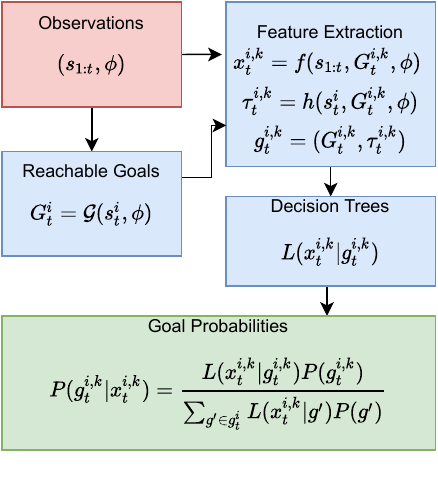}
    \caption{Diagram of the overall GRIT inference system.}
    \label{fig:grit}
    \vspace{-1.0em}
\end{figure}

An overview of GRIT's inference process is shown in Figure \ref{fig:grit}. As input during inference, GRIT takes the past observed trajectories of local vehicles $s_{1:t}$ and static scene information $\phi$. As output GRIT gives a probability distribution over possible goals for a vehicle. The first step in the inference process is to generate a set of possible goals for the vehicle. Next, a feature vector is extracted for each goal based on the past trajectories of all observed vehicles and static scene information. Decision trees are then used to infer a likelihood for each goal, before inferring a posterior probability distribution over goals via Eq. \eqref{goal_prob_overview}.

\subsection{Goal Generation}

In order to perform goal recognition for vehicle $i$ at time $t$, we first generate a set of possible goals $G^i_t = \mathcal{G}(s^i_t, \phi)$ from the vehicle's current state and static scene information. We assume a module $\mathcal{G}$ which gives us the possible goals. For example, goals could be extracted heuristically using road layout and vehicle state, as we do in our experiments.

\subsection{Feature Extraction}

For each vehicle $i$, for each possible goal $G^{i,k}_t$ at time $t$ we extract a feature vector $x^{i,k}_t=f(s_{1:t}, G^{i,k}_t, \phi)$ which will be used by the decision trees. These features can have binary or scalar values. We extracted the following features for each vehicle $i$ at each time $t$, which were chosen to be easily interpretable: Length of path to goal $\in \mathbb{R}^+_0$; In correct lane for goal $\in \{0,1\}$; Current speed $\in \mathbb{R}^+_0$; Current acceleration $\in \mathbb{R}$; Angle in lane $\in [-\pi, \pi)$; Distance to vehicle in front  $\in \mathbb{R}^+_0$; Speed of vehicle in front $\in \mathbb{R}^+_0$; Oncoming vehicle distance $\in \mathbb{R}^+_0$; Speed of oncoming vehicle $\in \mathbb{R}^+_0$.

\subsection{Goal Types}

Depending on a vehicle's position on the road relative to a goal location, the actions that vehicle must take to reach that goal can be quite different. For example, consider a vehicle with goal G1 in the scenario shown in Figure~\ref{fig:heckstrasse}. If the vehicle is approaching from the west, then it simply needs to continue straight on to reach its goal. However, if the vehicle is coming from the north, it needs to enter the T-junction and cross several lanes of traffic. We address this by considering a set of goal types, such as \textit{straight\_on} or \textit{turn\_left}. We could in principle train one tree for each goal that handles all goal types, but this would make the tree more complicated and thus less interpretable. We split the model up into separate trees for the different goal types to reduce the complexity of the model and improve interpretability. For each vehicle $i$ at time $t$, each possible goal location is assigned with a goal type $\tau^{i,k}_t = h(s^i_t, G^{i,k}_t, \phi)$ from the set $\mathcal{T}=\{\textit{straight\_on}, \textit{turn\_left}, \textit{turn\_right}, \textit{u\_turn\}}$. We assume these goal types are automatically assigned by the goal generation module.

\subsection{Decision Trees}

For each goal/goal type pair $g^{i,k}_t=(G^{i,k}_t, \tau^{i,k}_t)$, we train a decision tree which takes the feature values as input and outputs the likelihood $L(x^{i,k}_t|g^{i,k}_t)$ of the features given the goal and goal type. These likelihoods are combined with priors $P(g)$ to obtain a categorical posterior distribution over goals, as shown in Equation~\eqref{goal_prob_overview}. To obtain the output likelihood of the decision tree, we traverse the tree starting from the root based on the decision rule at each node until a leaf is reached. As shown in Figure \ref{fig:trained-tree}, each edge in the tree is assigned a weight. The likelihood value at each leaf node is calculated from the product of the initial likelihood of 0.5 with the weights on each edge leading to that leaf.

\subsection{Decision Tree Training}

We train each decision tree using the CART algorithm \cite{breiman_classification_1984}. When using CART, decision trees are expanded in an iterative manner starting from the root, greedily choosing the decision rule which maximises a certain criterion, in our case information gain. Cost complexity pruning \cite{hastie_elements_2009} is used to regularise the trees, and to aid interpretability the depth of the trees is limited.

Each decision tree is trained using the set of sampled vehicle states from the training set for which the relevant goal $G$ is reachable, while having the relevant goal type. These make up the dataset $D=\{(x_1, y_1), ..., (x_N, y_N)\}$, where $x_j$ is the set of features, and $y_j$ is the corresponding ground truth goal. A likelihood value is assigned to each node based on several sample counts: the total number of samples with goal $G$, $N_{G}=|\{j | y_j = G \}|$, the number of samples without goal $G$, $N_{\bar{G}}=N-N_G$, the number of samples at node $n$ with goal $G$, $N_{nG}=|\{j | x_j \in R_n \wedge\ y_j = G \}|$, and the number of samples at node $n$ without goal $G$, $N_{n \bar{G}}=N_n - N_{nG}$. Each of these counts is regularised using additive (Laplace) smoothing with hyperparameter $\alpha$. In many cases there is an imbalance of samples between $N_{nG}$ and $N_{n\bar{G}}$. To correct for this, we weight samples using the weights $w_G=N/N_G$ and $w_{\bar{g}}=N/N_{\bar{G}}$ so that the total weight given to samples with true goal $G$ and $\bar{G}$ is equal. The likelihood assigned to node $n$ of the tree is then given by:
\begin{equation} \label{likelihood}
    L_n=\frac{w_G N_{nG}}{w_G N_{nG}+w_{\bar{G}} N_{n\bar{G}}}
\end{equation}

\subsection{Verification} \label{verification_method}

One limitation of current prediction methods is the inability to guarantee safety through formal verification. GRIT can easily be verified due to the computational simplicity of decision tree inference, and the tree representation, which can be mapped into propositional logic. For example, we can verify that under certain conditions, certain nonsensical predictions will not be made. In order to perform verification, we first represented the model using propositional logic, and then verify a proposition $\Psi$ by proving that $\neg \Psi$ is unsatisfiable. We used the Z3 SMT solver \cite{de_moura_z3_2008} to perform the verification. In the event that verification of $\Psi$ fails, the solver provides a counterexample which can be useful to understand why the model makes certain predictions.

The decision trees can be represented using propositional logic with equality/inequality constraints by taking the conjunction of the statements given below. A Boolean variable $N_n$ is created for each node in the decision tree. The value of the variable is true if the node is reached and false otherwise. The variable for root nodes $N_{root}$ is always true. The decision rule at node $n$ is $D_n$. $D_n$ can represent a Boolean feature $x_j$ directly, or an inequality constraint on a scalar feature $c_n > x_j$. For each non-leaf node n, the child node variables follow the constraints $N_{n true child}=N_n \wedge D_n$, and $N_{n false child}=N_n \wedge \neg D_n$. If a leaf node is reached, then the likelihood output by the tree for goal/goal type pair $g=(G,\tau)$ is the likelihood $L_n$ at that node:
\begin{equation} \label{eqn:leaf}
    N_{leaf} \implies (L(x|g) = L_{leaf})
\end{equation}
\begin{equation} \label{eqn:posterior}
    P(g|x) = \frac{L(x|g) P(g)}{\sum_{g^\prime}{L(x|g^\prime) P(g^\prime)}}
\end{equation}
Some of the feature values can differ for different goals, such as ``in correct lane" and ``path to goal length". However, other features such current speed and acceleration are constrained to be the same regardless of the goal.

\section{Evaluation}

We evaluated GRIT and several baselines in four scenarios from two vehicle trajectory datasets. We show that: (1) GRIT has similar or better accuracy than the baselines; (2) GRIT inference is fast enough to run in real time; (3) the GRIT inference process is interpretable by humans; (4) properties of GRIT inference can be formally verified. A video showing GRIT is available at:  \url{https://www.five.ai/grit}.

\subsection{Datasets}

We evaluate GRIT and the baselines in the inD dataset \cite{bock_ind_2019} and the rounD dataset \cite{krajewski_round_2020}. Both of these datasets consist of vehicle trajectories recorded at several different junctions and roundabouts, along with local road layout maps which are provided in the Lanelet2 \cite{poggenhans_lanelet2_2018} format. 

\begin{figure}[t]
    \centering
    \includegraphics[width=0.485\textwidth]{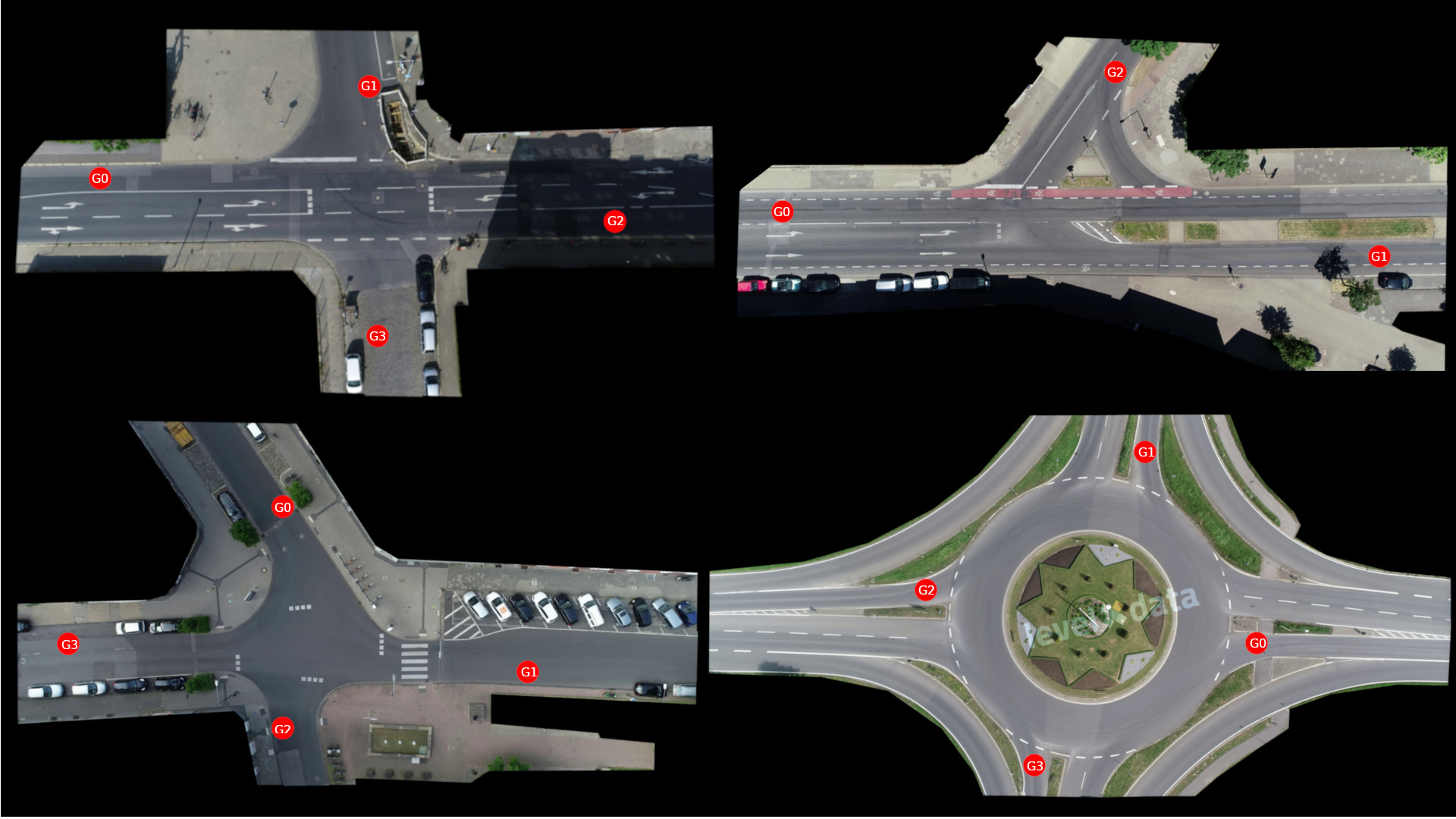}
    \caption{All scenarios used from the inD \cite{bock_ind_2019} and rounD \cite{krajewski_round_2020} datasets. Frankenberg (top left), Heckstrasse (top right), Bendplatz (bottom left) and Neuweiler (bottom right). The red dots show goal locations.}
    \label{fig:datasets}
\end{figure}

We trained and evaluated the models on three scenarios from the inD dataset, shown in Figure \ref{fig:datasets}. These included  ``Heckstrasse", a T-junction; ``Bendplatz", a marked crossroad with separate lanes for exiting; and ``Frankenberg", an unmarked crossroad \cite{bock_ind_2019}. We used one roundabout scenario ``Neuweiler" from the rounD dataset \cite{krajewski_round_2020}. Each scenario had a number of continuous recordings, with a typical duration of 20 minutes. For each scenario in the inD dataset, we randomly selected one recording for testing, one recording for validation (used for hyperparameter selection), and used the rest of the recordings for training. Due to the larger number of recordings in the rounD dataset, two recordings were randomly selected for validation and testing. The same split was used for all tested methods. Further information about data preprocessing is detailed in Appendix \ref{apx:preprocessing}.

\subsection{Baselines}

\subsubsection{GRIT-no-DT}
As a first baseline, we have included an ablation of GRIT in which the decision trees have been removed. This amounts to generating the set of possible goals based on the current vehicle state and road layout, and then re-normalising the prior probabilities for these goals to obtain the posterior goal distribution. In some scenarios, the prior probability of some goals is much higher than others, and simply always predicting the most common goal could achieve a high accuracy. This baseline gives context to the results achieved by the other methods by acting as a floor showing what a very simple method can achieve. 

\subsubsection{IGP2} In contrast to the learning based approaches, we also included the inverse planning based goal recognition used as part of IGP2 \cite{albrecht_interpretable_2020}. This method finds the optimal plan to each possible goal from both the vehicle's current position, and first observed position. The goal likelihood is then calculated based on the cost difference between these two plans, with a larger difference leading to lower likelihood. Similarly to GRIT, a Bayesian posterior probability distribution over goals is then calculated. The predictions made by IGP2 are highly interpretable, however the planning process used by IGP2 is computationally complex which makes inference slow and verification infeasible. The implementation details for IGP2 are given in Appendix \ref{apx:igp2}.

\subsubsection{LSTM} As another baseline, we trained a recurrent neural network architecture based on Long Short-Term Memory (LSTM) \cite{lstm} for each scenario individually to directly predict goal probabilities. The model architecture and training setup is detailed in Appendix \ref{apx:lstm}.

\subsubsection{GR-ESP} We implemented a goal recognition method based on trajectories sampled from the multi-agent deep generative model (called ESP) of the PRECOG system \cite{rhinehart_precog_2019}. We call this baseline Goal Recognition with ESP (GR-ESP). PRECOG is a deep learning based model which was shown to make robust, goal-aware planning decisions when conditioned on goal-positions and was able to accurately estimate the likeliest future trajectory an agent could take to safely reach a goal. Given $K$ trajectory samples drawn from ESP for a vehicle, we define the probability of a goal $G$ as the normalised count of trajectories whose endpoints are closest to $G$. The generated trajectories of ESP are fixed-length, therefore we perform repeated sampling to obtain trajectories with a suitable length. Exact training and implementation details are shown in Appendix \ref{apx:gresp}.

\subsection{GRIT Implementation}

We manually annotated each scenario with goal locations and define the possible goals $\mathcal{G}(s^i_t, \phi)$ for vehicle $i$ at time $t$ as the set of goal locations reachable from $s^i_t$ under traffic rules allowed by the local road layout $\phi$. Reachability checking is performed using functionality built into the Lanelet2 library \cite{poggenhans_lanelet2_2018} which uses Dijkstra's shortest path algorithm.

The prior probabilities for each goal/goal type pair were estimated from their frequency in the training dataset, with Laplace smoothing. The maximum depth of the decision trees was limited to 7. The parameters for Laplace smoothing and cost complexity pruning \cite{hastie_elements_2009} were chosen by grid search, with a separate set of parameters selected for each scenario.

\subsection{Accuracy and Entropy}

One evaluation metric used was {\bf accuracy} -- the fraction of test samples for which the true goal was assigned the highest probability. Another metric which was examined is {\bf normalised entropy}, which is the entropy of the posterior goal distribution divided by the entropy of a uniform distribution. This gives a measure of the uncertainty of the model about the vehicle's goal.

\begin{figure}[t]
\vspace{-0.4cm}
\begin{tabular}{cc}

\hspace{-0.1cm}\subfloat{\includegraphics[width = 1.6in]{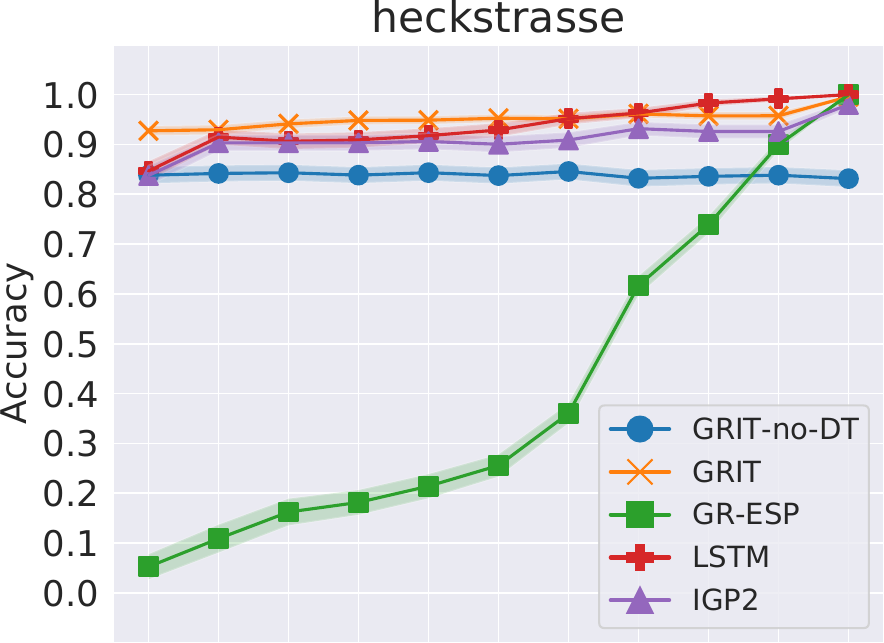}} &
\hspace{-0.1cm}\subfloat{\includegraphics[width = 1.6in]{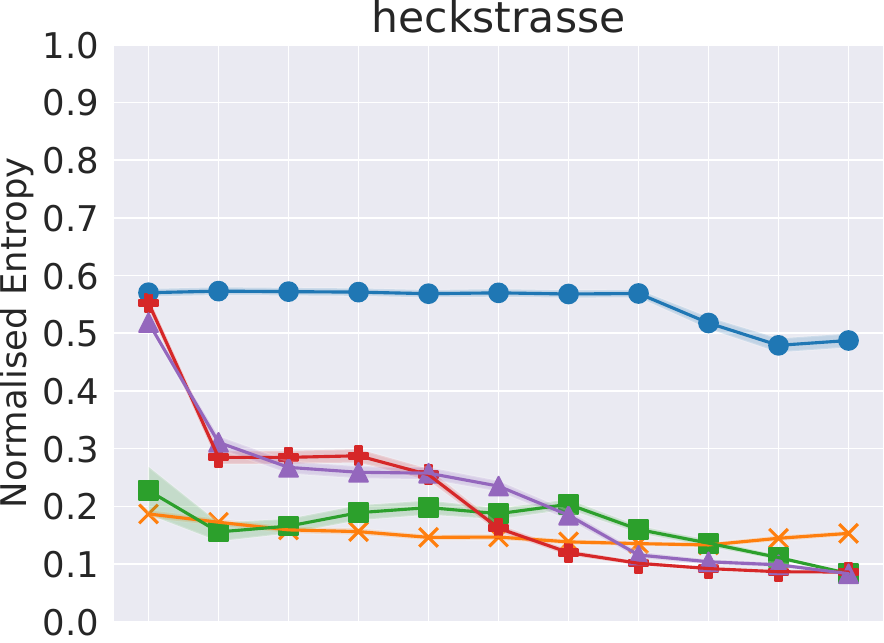}} \vspace{-0.4cm}\\
\vspace{-0.4cm}
\hspace{-0.1cm}\subfloat{\includegraphics[width = 1.6in]{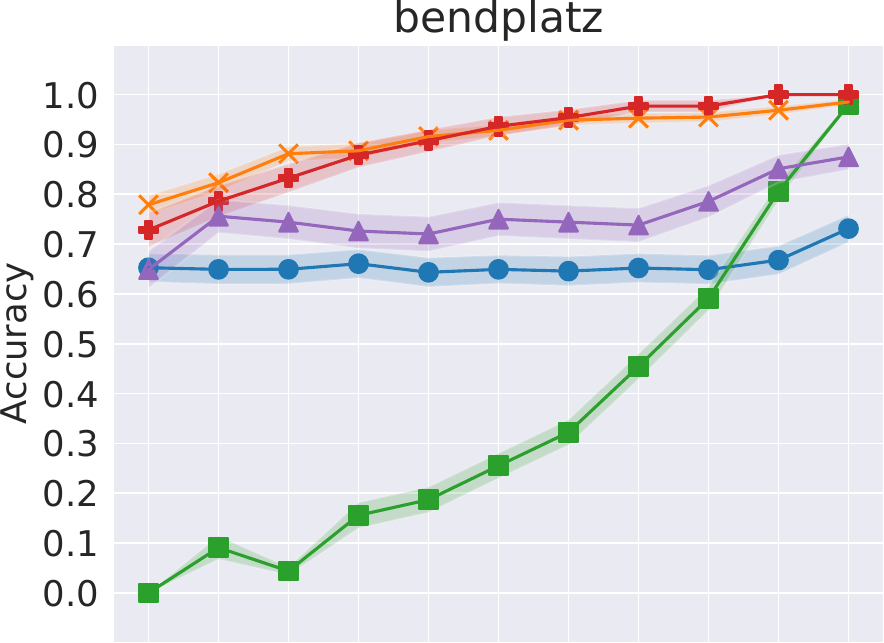}} &
\hspace{-0.1cm}\subfloat{\includegraphics[width = 1.6in]{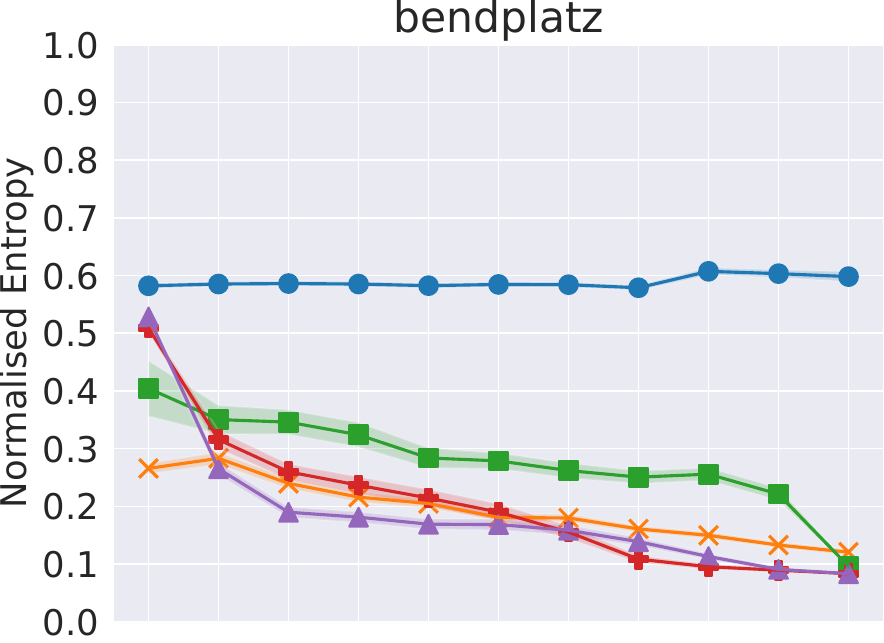}}\\
\vspace{-0.4cm}
\hspace{-0.1cm}\subfloat{\includegraphics[width = 1.6in]{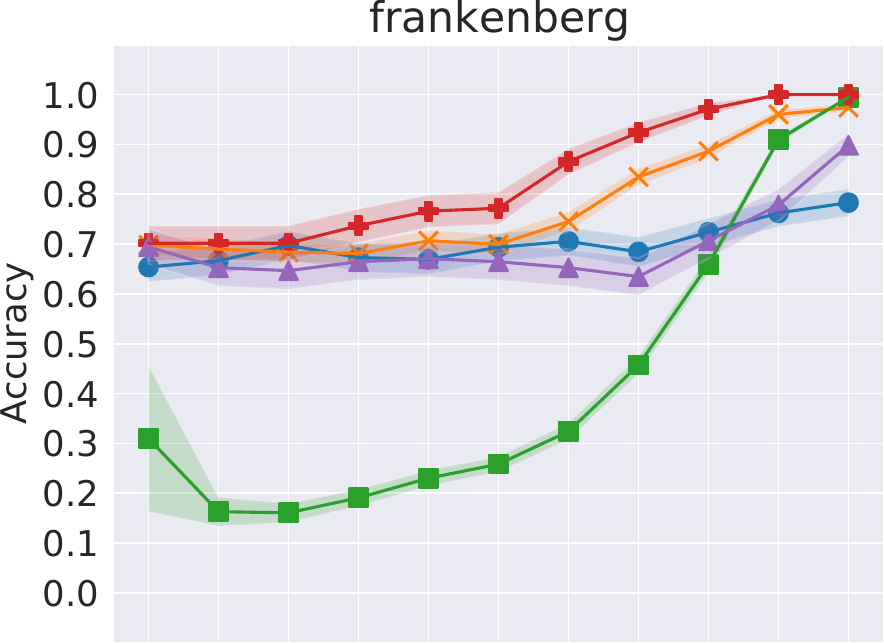}} &
\hspace{-0.1cm}\subfloat{\includegraphics[width = 1.6in]{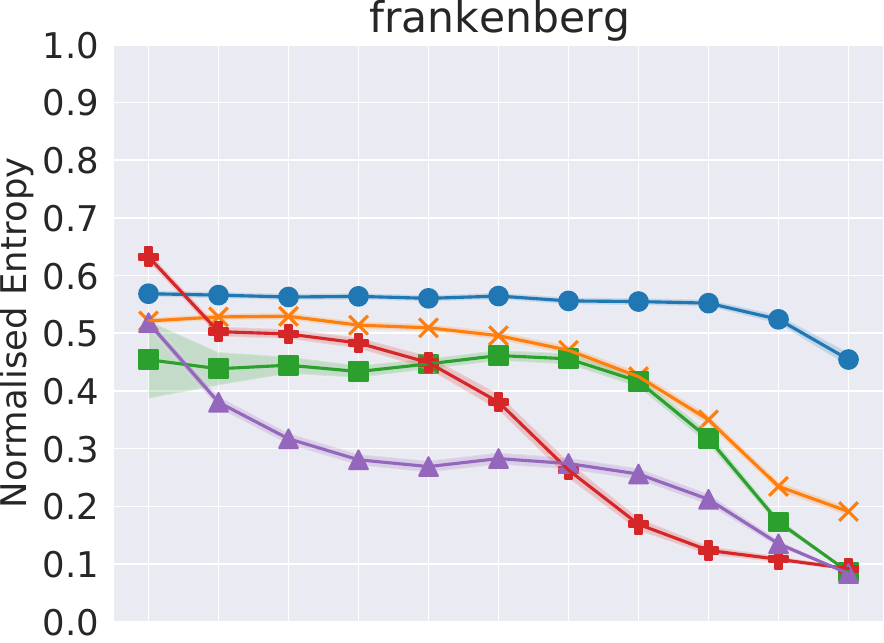}}\\
\hspace{-0.1cm}\subfloat{\includegraphics[width = 1.6in]{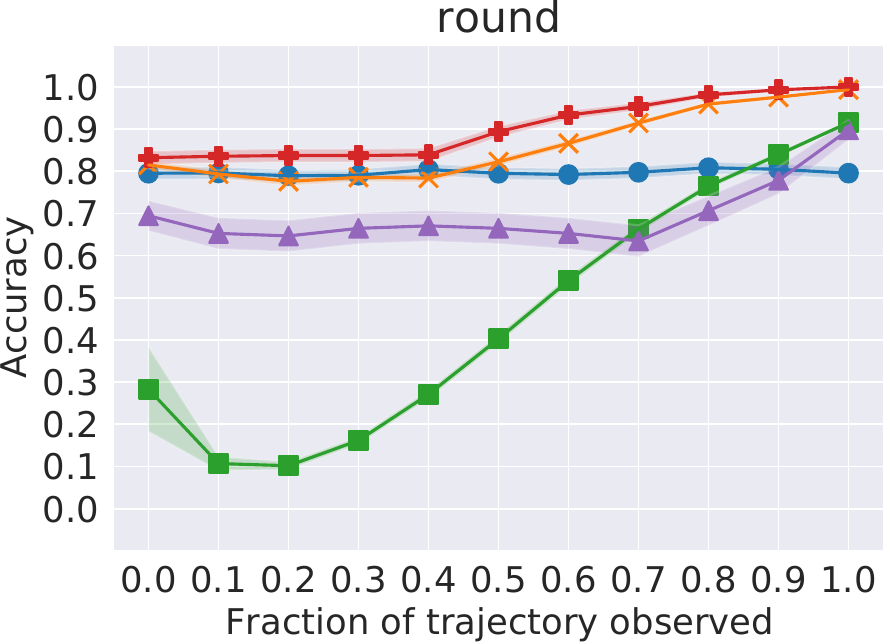}} &
\hspace{-0.1cm}\subfloat{\includegraphics[width = 1.6in]{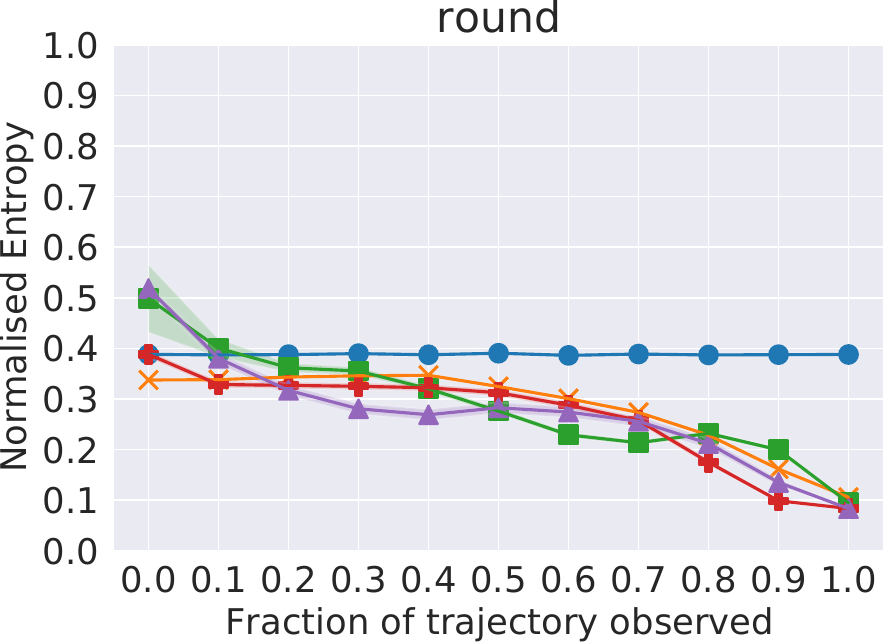}}
\end{tabular}
\caption{Goal recognition accuracy (left) and normalised entropy (right) for each method on each scenario. The shaded areas show standard error.}
\label{fig:fraction-observed}
\end{figure}

The evolution of accuracy and entropy as the fraction of the trajectory observed increases is shown in Figure \ref{fig:fraction-observed}. Across all methods except GRIT-no-DT, accuracy increases as the fraction of the trajectory observed increases. Similarly, normalised entropy tends to decrease as more of the trajectory is observed, showing how the models become more certain about a vehicle's goal as more observations are made. The LSTM model achieved the highest accuracy overall. In the Heckstrasse and Bendplatz scenarios, GRIT achieved a similar accuracy to the LSTM, however GRIT had slightly lower accuracy than the LSTM in Frankenberg and rounD. The LSTM model could be extracting more information from the raw trajectory than is represented in the tree features used by GRIT, leading to higher accuracy than GRIT in some cases. As expected, the accuracy of GRIT-no-DT is lower than that of full GRIT and LSTM, as it does not have access to observations other than the set of reachable goals. IGP2 also achieves lower accuracy than GRIT and LSTM. One reason for this is that the inverse planning used in IGP2 sometimes fails to find a plan to the true goal, in which case this goal is given zero probabability \cite{albrecht_interpretable_2021}. The average fraction of samples over time for which no plan was found to the true goal in each scenario was 0.013 for Heckstrasse, 0.070 for Frankenberg, 0.101 for Bendplatz, and 0.010 for Round. Another reason for the lower accuracy could be that the goal priors used for IGP2 are less fine grained than those used for GRIT, as IGP2 uses a prior probability for each goal, rather than each goal/goal type pair. GR-ESP achieves significantly lower accuracy other methods, except when the fraction of trajectory observed is very close to one.

\subsection{Inference Time}

\begin{figure}[t]
    \centering
    \includegraphics[width=2.0in]{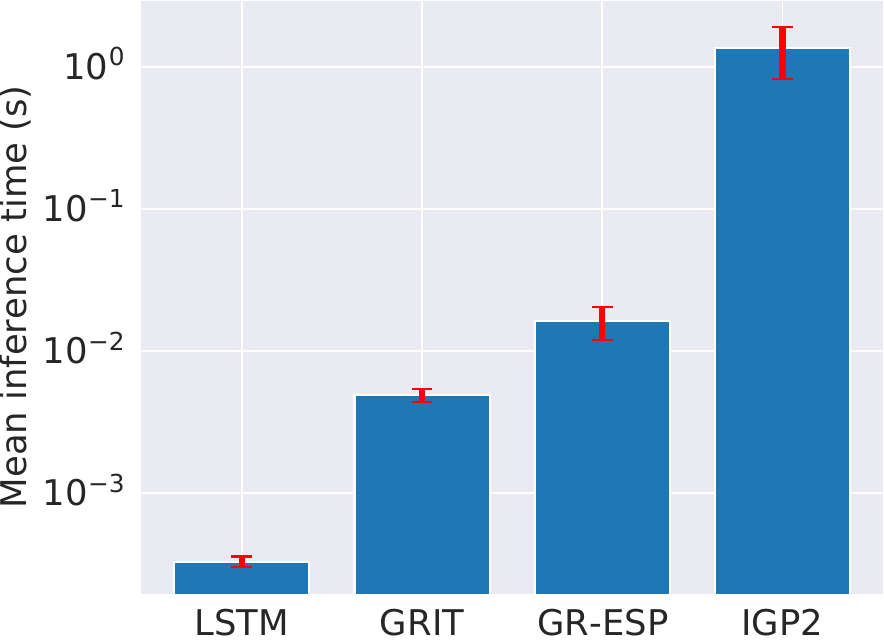}
    \caption{Mean inference time per vehicle in seconds on log-scale with standard error.}
    \label{fig:inference_time}
\end{figure}

A comparison of mean inference times for each method is shown in Figure \ref{fig:inference_time}. All methods other than IGP2 are fast enough to run in real time. The fastest method is LSTM, followed by GRIT. The majority of time during GRIT inferences is taken up by the feature extraction. GR-ESP is slower than LSTM and GRIT, due to its repeated sampling process. IGP2 is by far the slowest method due to its computationally intensive planning process.

\subsection{Interpretability}

\begin{figure}[t]
    \centering
    \includegraphics[height=0.40\textheight]{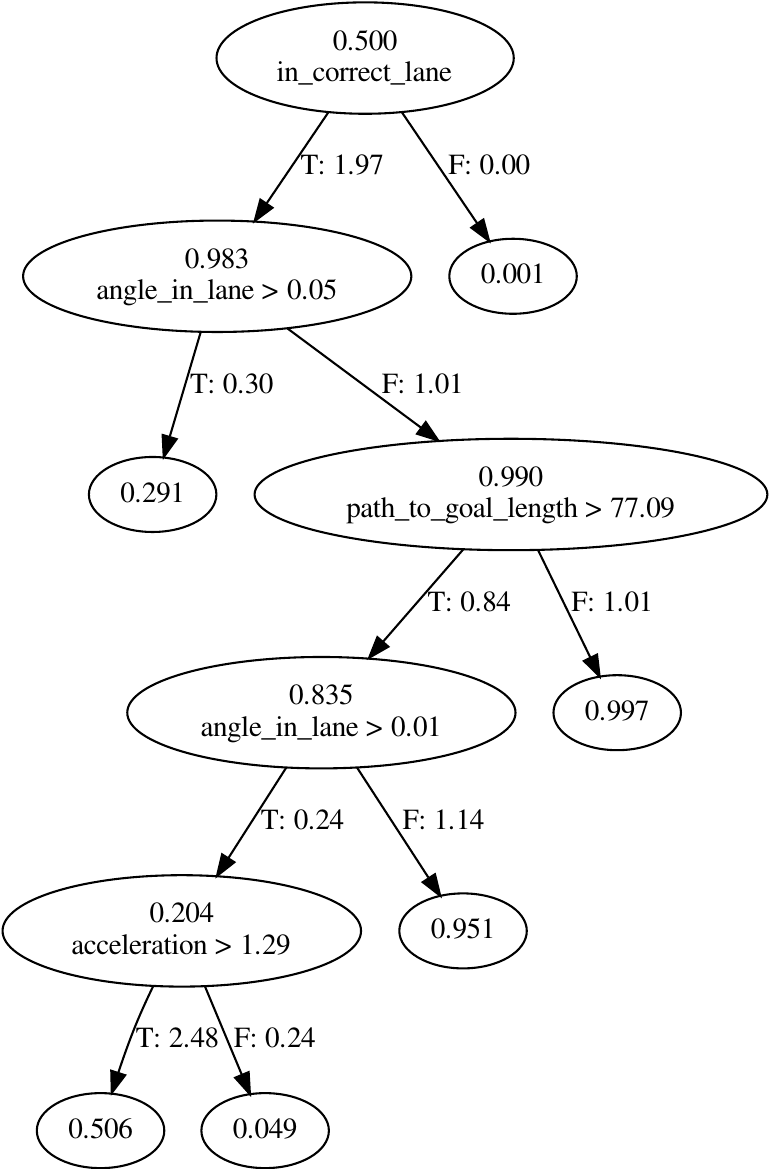}
    \caption{Trained decision tree for the Heckstrasse scenario goal G1, goal type \textit{straight\_on}. Multiplicative weights are assigned to each True (T) and False (F) edge. The first value shown in each node is the cumulative product of the initial likelihood 0.5 and the weights of edges traversed so far, representing the likelihood $L(x|g)$. Beneath this, the decision rule for each node is shown.}
     \label{fig:trained-tree} 
    \vspace{-1.0em}
\end{figure}

We found that the trees learned by GRIT were human-interpretable, with average depth of 6.19. For example, take the trained decision tree shown in Figure \ref{fig:trained-tree}. If the  top-left leaf node with likelihood 0.291 is reached, an explanation with weights for each factor can easily be extracted: ``Goal G1 \textit{straight\_on} has a likelihood of 0.291 because the vehicle is in the correct lane (weight 1.97) and the vehicle's angle in lane is greater than 0.05 radians to the left (weight 0.30)". Examining the scenario in Figure \ref{fig:heckstrasse} shows that this interpretation makes sense. Although a vehicle which reached the 0.291 likelihood leaf node in Figure \ref{fig:trained-tree} is in the correct lane to go straight on, the fact that it is angled to the left suggests that it will turn left rather than going straight on, leading to a low likelihood for the \textit{straight\_on} goal.

\subsection{Verification}
\label{verification}

Using the method described in Section \ref{verification_method}, we were able to verify several properties of our learned trees. In failed verification attempts, the method provided a counter-example to explain why our original intuition was incorrect, and this knowledge could be useful for improving the models. Here we give examples for the Heckstrasse scenario, shown in Figure \ref{fig:heckstrasse}. We were also able to prove similar properties in other scenarios. Such verification is currently not possible with the deep learning models -- we never quite know what these methods will predict. The verification process was relatively fast, taking an average of 65.2 milliseconds to verify a proposition for each tree. Formal definitions of each proposition are given in Appendix \ref{apx:verify}.

\subsubsection{Predict goal corresponding to lane} As can be seen from Figure \ref{fig:heckstrasse}, if a vehicle is coming from the west, there are two possible lanes: one marked as the lane for exiting left, and one marked as the lane for continuing straight on. One reasonable expectation of a goal recognition model would be that if a vehicle is in the correct lane for a goal, then that goal will be assigned the highest probability. We successfully verified the proposition ``If the vehicle is in the correct lane for G2 (\textit{turn\_left}), then G2 is assigned the highest probability". Verification failed for the equivalent proposition for G1. However, the solver provided the feature values shown in Table \ref{counterexample} as a counterexample, which can still teach us about the way in which the model works. Despite the fact that the vehicle is in the lane for G1 (\textit{straight\_on}), the vehicle is angled to the left in its lane, while going at a slow speed and decelerating, and is still far from the junction entry (path\_to\_goal\_length 78.09 meters). This together is reasonable evidence towards G2 being the true goal, and in such a case assigning the higher probability to G2 is correct. We also attempted to verify a relaxed version of this proposition, in which we verify that the probability of G1 is always above a certain lower bound. We successfully verified that if the vehicle is in the correct lane for G1, then the probability assigned to G1 is always greater than 0.2. 

\subsubsection{Goal distribution entropy} As a vehicle travels closer towards its goal, a reasonable expectation is that we should become more certain about what its goal is -- that is, the entropy of the distribution over goals should decrease, or at least stay constant. If there are just two possible goals, then it is equivalent to show that if one goal has higher probability than the other, then the probability of the most probable goal will not decrease as the length of the path to the goal decreases. We attempted to verify this for the situation where a vehicle is approaching from the east in the Heckstrasse scenario. However, in this case the verification failed. If a vehicle is in a state such as that shown in Table~\ref{counterexample}, the model predicts with high certainty that a vehicle is going to turn at the junction. The prediction is biased towards the goal G2, \textit{turn\_left} due to the vehicle's angle in the lane, although the vehicle is still quite distant from the junction entry. However, if the vehicle continues further along the road and has still not switched lane, the uncertainty over goals can actually increase. In this situation it makes sense for uncertainty to increase, because the vehicle took actions that were irrational for the goal it originally started moving towards, showing that our original intuition was incorrect.

\begin{table}[t]
\centering
\caption{Generated counterexample to the proposition: ``If the vehicle is in the correct lane for G1 (\textit{straight\_on}), then G1 is assigned the highest probability". Despite the fact that the vehicle is in the lane for G1 (\textit{straight\_on}), there is some evidence of G2 being the true goal: the vehicle is angled to the left in its lane, while going at a slow speed and decelerating, and is still far from the junction entry.}
\label{counterexample}
\begin{tabular}{lll}
\textbf{Features}             & \textbf{Goal 1} & \textbf{Goal 2} \\ \hline
path\_to\_goal\_length    & 78.09      & 57.09      \\
in\_correct\_lane         & True       & False      \\
speed                     & 0          & 0          \\
acceleration              & -1         & -1         \\
angle\_in\_lane           & 0.03125    & 0.03125    \\
vehicle\_in\_front\_dist  & 32.87      & 32.87      \\
vehicle\_in\_front\_speed & 0          & 0          \\
oncoming\_vehicle\_dist   & None       & None       \\
goal likelihood           & 0.04874    & 0.9242     \\
goal probability          & 0.2014     & 0.7985  
\end{tabular}
\end{table}

\subsubsection{Verification across all scenarios} The verification cases mentioned above apply to specific situations in a scenario, however it is also possible to verify some propositions more broadly across all scenarios. One such proposition is that changing a single input feature while leaving all other features unchanged will have a certain effect on the goal likelihood. More specifically, we ran verification for the proposition ``If a vehicle is in the correct lane for a goal then that goal should have the same or higher likelihood than if the vehicle is not in the correct lane, if all other features remain unchanged". In total there were 47 goal/goal type pair across all scenarios, each having a separate decision tree learned by GRIT. Verification was successful for all but 4 of these goal/goal type pairs. Upon inspection of the counterexamples generated in those cases, the models were still giving reasonable likelihoods given the features. For example, in one counterexample a vehicle is in the incorrect lane to turn, but is travelling at a low speed. In such a situation the vehicle could have slowed down in order to turn, so assigning a high likelihood to turning makes sense.

\subsubsection{Stopping for oncoming vehicles} It is also possible to verify predictions made by the model in more complicated situations, such as stopping for oncoming vehicles. Verification was performed across all scenarios for the proposition  ``If a vehicle (V1) has stopped, and there is no stopped vehicle in front of V1, and there is an oncoming vehicle (V2) in a lane which V1 must cross to reach certain goal, then that goal will have the same or higher likelihood for V1 than if there was no oncoming vehicle V2, all other features being unchanged". For this proposition, verification was successful for 44/47 of the total goal/goal type pairs.

\section{CONCLUSIONS}

We presented GRIT, a goal recognition method for autonomous vehicles which makes use of decision trees trained from vehicle trajectories. We have shown empirically in four scenarios from two vehicle trajectory datasets that GRIT achieves high goal recognition accuracy and fast inference times, and that the learned tree models are both interpretable and verifiable. To the best of our knowledge, GRIT is the first verifiable goal recognition method for autonomous vehicles.

In this work we trained and tested GRIT on several specific ``fixed-frame'' scenarios. Future work could extend GRIT for open-world driving, by training one decision tree for each goal type across scenarios and dynamically generating possible goal locations.
Recent work \cite{liu_improving_2019} has shown that decision trees trained by distilling knowledge from deep neural networks can achieve higher accuracy than those trained from scratch. To further improve the accuracy of GRIT, future work could investigate knowledge distillation from deep neural networks to decision trees. Another future extension of GRIT could be to handle occlusion \cite{hanna_interpretable_2021}.

\section*{ACKNOWLEDGMENT}

This research was in part supported by the Royal Society via an Industry Fellowship (S. V. Albrecht).

\bibliographystyle{IEEEtran}
\bibliography{library.bib}

\appendix
\label{apx}
\subsection{Data Preprocessing}
\label{apx:preprocessing}
We manually annotated each of the scenarios with goal locations, as can be seen in Figure \ref{fig:heckstrasse}. These included junction/roundabout exits and visible lane ends. We determined the ground truth goal of each vehicle by finding the first goal for which the trajectory passes within a defined distance (1.5m in the Heckstrasse example). Vehicle trajectories for which none of the predefined goals were reached, were discarded. To create training data and test data for GRIT, each trajectory was first trimmed up to the point where the goal was reached. Following this 11 evenly timed samples were taken from each trajectory. Each sample contained the state history of the vehicle of interest up to that point in time, along with the state history of other vehicles from the point at which the vehicle of interest was first observed.

\subsection{Baseline Implementations}
\subsubsection{LSTM}
\label{apx:lstm}
The input to the LSTM is a raw state sequence $s^i_{1:t}$ of vehicle $i$ and the target for each time step is the true goal $G^i_t$. Our model architecture is built from a single-layer LSTM. The hidden unit of each cell has size 64. The outputs of the LSTM at each time step is pushed through a fully connected (FC) network with one hidden layer of dimension 725 and ReLU activation. The weights of the FC layer is initialised using normally distributed Glorot initialisation \cite{pmlr-v9-glorot10a}. We minimise the mean total loss cross-entropy using the Adam optimiser \cite{diedrik2015adam} with a learning rate of $5\times10^{-4}$. We train for 1,000 epochs with early stopping and using a batch size of 10 trajectories. Our FC layer is regularised with dropout with $p=0.2$. We schedule the learning rate to decrease by a factor of 0.5 if the best validation loss did not improve for 10 consecutive epochs.

\subsubsection{GR-ESP}
\label{apx:gresp}
GR-ESP predicts from $t=0$ the joint future state of length $T$ (4 seconds) of all vehicles given a tuple $\{s_{-\tau:0}, \chi\}$. The first element is the past joint state with length $\tau$, and $\chi\in \mathbb{R}^{w\times h \times 8}$ is composed of feature-maps that represent the road surface, road markings, and vehicles at $t=0$. We downsample our data set to 10 fps, then discretise it to time-steps using a moving window with a step-size of 0.5 seconds (5 frames) and a length of $\tau+T$ seconds, in our experiments $\tau=2$. Trajectories in the window that are shorter than $\tau+T$ seconds are padded using a straight-ahead constant-velocity assumption. We train GR-ESP for 40,000 steps using the original hyper-parameters and sample $K=100$ trajectories per time-step. During testing, if no goals are reached on the first sampling, then we repeatedly generate new trajectories up to $R$ additional times. At each repeated generation we condition on the final $\tau$ seconds of the previously generated future trajectory. We set $R=2$ for the rounD scenario and $R=1$ otherwise.

\subsubsection{IGP2}
\label{apx:igp2}
The goal recognition module of IGP2 is implemented as described in \cite{albrecht_interpretable_2020}, using all macro-actions and maneuvers to predict trajectories, except \textit{Stop}. We run the velocity smoothing module with parameters $\lambda = 10$, timestep $\Delta t = 0.1$, $a_{max} = 5$ and $v_{max}$ set to the episode speed limit. To improve convergence during velocity smoothing, we are using the objective function:
\begin{equation} \label{eq:smooth}
\begin{aligned}
& &\min_{x_{2:n}, v_{2:n}} & & & \sum_{t=1}^n (v_t - \kappa(x_t))^2 +  \lambda\sum_{t=1}^{n-1} (v_{t+1} - v_t)^2 \\
\end{aligned}
\end{equation}

The reward terms are normalised between 0 and 1 according to their distributions across both datasets, with values falling beyond three standard deviations of the distribution being clipped. The reward weights and the free parameters of the macro-actions and maneuvers are reported in table \ref{IGP2_parameters}.

\begin{table}[]
\centering
\caption{IGP2 reward weights and free parameters of macro-actions and maneuvers used for each dataset.}
\label{IGP2_parameters}
\begin{tabular}{lll}
\textbf{Parameters}                 & \textbf{InD dataset} & \textbf{RounD dataset} \\ \hline
time to goal reward weight              & 0      & 0.01      \\
angular velocity reward weight          & 0        & 0.01      \\
heading reward weight                   & 1000      & 10         \\
acceleration reward weight              & 0      & 0.01      \\
give way distance                       & 10         & 10          \\
give way lane angle threshold           & $\pi/6$    & $\pi/6$          \\
give way turn target threshold          & 1          & 1    \\
maneuver point spacing                  & 0.25      & 0.25      \\
maneuver max speed                      & speed limit         & speed limit \\
maneuver min speed                      & 3         & 3       \\
switch lane target switch length        & 20        & 10     \\
switch lane minimum switch length       & 5         & 5  
\end{tabular}
\end{table}

\subsection{Experiment Hardware Specifications}
All experiments were carried out on a server with two AMD EPYC 7502 CPUs and eight Nvidia GeForce RTX 2080 Ti GPUs. The GPUs were used for the neural networks in both the LSTM and GR-ESP baselines, and all other computation was performed on the CPUs.

\subsection{Verification}
\label{apx:verify}

In this section, formal definitions are be given for each of the verification propositions mentioned in Section \ref{verification}. The entire set of features are represented by $x^g_t$, where $g$ is an identifier for a goal, and $t$ is a certain timestep or instance of the scene. Individual feature values are represented by $x^{g,f}_t$, where $f \in \mathcal{F}$ is an identifier for a feature. The identifiers used are: path to goal length: $path$, in correct lane: $l$, speed: $spd$, vehicle in front speed: $fs$, vehicle in front distance: $fd$, oncoming vehicle distance: $ond$. The proposition ``If there are two possible goals G1 and G2, and the vehicle is in the correct lane for G2 (\textit{turn\_left}), then G2 is assigned the highest probability" was represented as:
$$ x^{G1,l}_t \wedge \neg x^{G2,l}_t \implies P(G1|x^{G1}_t) > P(G2|x^{G2}_t)$$

The proposition ``If there are two possible goals, and one goal has higher probability than the other, then the probability of the most probable goal will increase or stay constant as the length of the path to the goal decreases, if all other features remain unchanged" was represented as:
$$\bigwedge_{g \in \{G1,G2\}} ( (x^{g,path}_{t1} > x^{g,path}_{t2} )
\bigwedge_{f \in \mathcal{F}, \atop f \neq path} (x^{g,f}_{t1}=x^{g,f}_{t2}))$$
$$\implies ((P(G1|x^{G1}_{t1}) < P(G2|x^{G2}_{t1}) \implies P(G2|x^{G2}_{t2}) $$
$$\geq P(G2|x^{G2}_{t1})) \wedge (P(G1|x^{G1}_{t1}) > P(G2|x^{G2}_{t1}) $$
$$\implies P(G1|x^{G1}_{t2}) \geq P(G1|x^{G1}_{t1})))$$

The proposition ``If a vehicle is in the correct lane for a goal then that goal should have the same or higher likelihood than if the vehicle is not in the correct lane, if all other features remain unchanged" was represented as:
$$x^{g,l}_{t1} \wedge \neg x^{g,l}_{t2} \bigwedge_{f \in \mathcal{F}, \atop f \neq lane} x^{g,f}_{t1}=x^{g,f}_{t2} \implies L(x^{g}_{t1}|g) \geq L(x^{g}_{t2}|g)$$

The proposition ``If a vehicle (V1) has stopped, and there is no stopped vehicle in front of V1, and there is an oncoming vehicle (V2) in a lane which V1 must cross to reach certain goal, then that goal will have the same or higher likelihood for V1 than if there was no oncoming vehicle V2, all other features being unchanged" was represented as:
$$x^{g,spd}_{t1} < 1 \wedge x^{g,fs}_{t1} = 20 \wedge x^{g,fd}_{t1} = 100 \wedge x^{g,ond}_{t1} = 100 \wedge $$
$$x^{g,ond}_{t2} = 20 \bigwedge_{f \in \mathcal{F}, \atop f \neq ond} x^{g,f}_{t1}=x^{g,f}_{t2} \implies L(x^{g}_{t2}|g) \geq L(x^{g}_{t1}|g)$$

\end{document}